\documentclass{IOS-Book-Article}

\usepackage{mathptmx}
\usepackage{soul}\setuldepth{article}
\usepackage[numbers]{natbib}
\usepackage{booktabs}
\usepackage{url}
\usepackage{microtype}
 
\def\hb{\hbox to 10.7 cm{}}

\begin{document}

\pagestyle{headings}
\def\thepage{}

\begin{frontmatter}              

\title{Evaluating Sentence Segmentation and Word Tokenization Systems on Estonian Web Texts}


\author{\fnms{Kairit} \snm{Sirts}%
\thanks{Corresponding Author: Kairit Sirts; E-mail:
kairit.sirts@ut.ee.}}
and
\author{\fnms{Kairit} \snm{Peekman}}

\address{Institute of Computer Science, University of Tartu, Estonia}

\begin{abstract}
	Texts obtained from web are noisy and do not necessarily follow the orthographic sentence and word boundary rules. Thus, sentence segmentation and word tokenization systems that have been developed on well-formed texts might not perform so well on unedited web texts. In this paper, we first describe the manual annotation of sentence boundaries of an Estonian web dataset and then present the evaluation results of three existing sentence segmentation and word tokenization systems on this corpus: EstNLTK, Stanza and UDPipe. While EstNLTK obtains the highest performance compared to other systems on sentence segmentation on this dataset, the sentence segmentation performance of Stanza and UDPipe remains well below the results obtained on the more well-formed Estonian UD test set.
\end{abstract}

\end{frontmatter}

\section{Introduction}

Sentence segmentation and word tokenization are the first pre-processing steps for most NLP tasks. Existing methods for addressing these steps are either rule-based or trained on annotated corpora. Rule-based systems for sentence and word tokenization make certain assumptions about the well-formedness of the text, for instance, that the sentences start with a capital letter and end with one of the valid sentence-final punctuation marks. Data-driven systems rely on the specifics of the annotated training corpora, most of which are also based on well-formed text genres like fiction or news articles.

Web texts, such as forum or blog posts and user commentaries, are typically unedited and thus the well-formedness of such texts can not be guaranteed. Some types of social media texts, tweets, for instance, have very distinct characteristics and thus, specific tools have been created for their token and sentence segmentation \cite{o2010tweetmotif}. Although some examples of tokenizers and sentence segmentation systems developed specifically for other types of web and social media texts exist \cite{proisl2016somajo}, it is generally assumed that standard tools are applicable to web texts as well.

For Estonian, both rule-based and model-based tokenizers and sentence segmentation systems are available. The rule-based system is part of the EstNLTK library \cite{orasmaa2016estnltk} that gathers existing NLP tools for Estonian. Data-driven solutions are available in several pre-trained parsing pipeline models such as Stanford Stanza \cite{qi2020stanza} and UDPipe \cite{straka2016udpipe}. Both of these systems have achieved good results on both tokenization and sentence segmentation on Universal Dependencies (UD) Estonian treebank test set as evaluated in the CoNLL 2018 Shared task on parsing raw text \cite{zeman2018conll}. 

In addition to the Estonian UD treebank consisting of standard texts, also a smaller Estonian Web Treebank (EWT) \cite{muischnek2019cg} is available, the tokenization and sentence segmentation of which are based on automatic analysis with EstNLTK. Thus, although this corpus can be used for evaluating the performance of NLP tools on web texts, the correctness of the token and sentence segmentation of this corpus cannot be guaranteed.

Our goal in this paper is to evaluate the existing sentence segmentation and word tokenization systems on Estonian web texts.
For that purpose, we first manually annotated the token and sentence boundaries of the EWT corpus. In the following, we will first describe the annotation process and then present the evaluation of token and sentence segmentation performance using both EstNLTK and the pretrained Stanza and UDPipe models.\footnote{The annotated dataset is available at: \url{https://github.com/ksirts/EWTB_sentence_seg}}

\section{Related Work}

Although sentence boundary detection is usually regarded as a solved problem or, at least, a problem that does not warrant much effort and attention, \cite{read2012sentence} showed that this is not the case. They performed experiments with all sentence segmentation systems available at the time on various English corpora and showed that results are far from perfect and vary depending on the domain of the corpus. Although at the time of conducting their study, models based on neural architectures were not available yet, it is highly likely that when adding the current existing neural models to the experimental plan of \cite{read2012sentence}, the overall picture would remain roughly the same.

Tweets constitute a very specific domain of social media texts and thus, various attempts have been made to develop sentence segmentation and word tokenization systems specifically for handling tweets \cite{o2010tweetmotif}. \cite{vcibej2016normalisation} describe an annotation effort to create a sentence and token boundary annotated tweet corpus in Slovene. They mention several annotation issues also relevant in our work, in particular, in relation to ambiguous sentence endings. For instance, depending on the context, multiple dots can be interpreted as a pause or as sentence-final punctuation.

Annotating and predicting sentence boundaries in social media texts is in many ways similar to handling sentence boundaries in speech transcriptions, especially when only textual features extracted from the speech transcripts are used \cite{tilk2016bidirectional}.
Several previous works that have studied the issues related to sentence boundary segmentation in speech transcripts have acknowledged the issue related to the question of what constitutes a sentence, i.e. what units have to be annotated. 
For instance, \cite{westpfahl2018syntax} describe a sentence boundary annotation effort on speech transcripts based on syntactic information, i.e. a sequence annotated as a sentence should form a syntactially complete unit.
As another example, \cite{stepikhov2013resolving} assessed the inter-annotator agreements of annotating sentence boundaries in Russian transcribed speech data. They found that when using a threshold of 60 \% for deciding the majority annotations, more than 70 \% of the annotations would have been lost, suggesting that the inter-annotator agreements were relatively low. Our task is simpler compared to these works in a sense that while we cannot assume that the writers of the web texts have followed conventional grammatical norms, they nevertheless have used various punctuation marks and emoticons that help to determine potential sentence boundaries.

\section{Corpus Annotation}
\label{sec:corpus}

As a part of this project, the texts included in the Estonian Web Treebank (EWT) \cite{muischnek2019cg} were manually annotated with sentence boundaries. The sentence boundary annotation involves determining where one sentence ends and the next one starts and, typically, this is decided based on orthographic sentence boundary markers. However, as the web texts are unedited and the writers might not adhere to usual orthographic rules, relying on common orthographic sentence end markers might, in some cases, result in overly long sentences. Thus, in the context of web texts, we propose to distinguish between two types of sentences:
\begin{enumerate}
    \item Orthographic sentences that typically start with a capital letter and end with one the commonly used sentence final punctuation marks (.?!);
    \item Syntactic sentences that might not adhere to the orthographic rules but that are syntactically complete, i.e. the syntactic head of each word resides inside the sentence. 
\end{enumerate}

Originally, our corpus did not contain paragraph boundaries. However, paragraph information, if available, can be very helpful for sentence segmentation systems as it gives a free sentence boundary at the end of every paragraph.
The EWT texts originate from the EtTenTen 2013 corpus \cite{muischnek2016eesti} which contains the paragraph information and thus, we reintroduced the paragraph boundaries back into the corpus. The final corpus consists of 32 documents and in total of 522 paragraphs.

We instructed the annotators to separately annotate the locations of both types of sentence boundaries. For instance, consider a paragraph that orthographically consists of a single long sentence, but syntactically contains several shorter syntactically independent parts, each separated from each other with a comma. The expected annotation of this paragraph was to add an orthographic sentence boundary at the end of the paragraph and syntactic sentence boundaries after each syntactically complete clause.

The corpus was annotated in two equal parts; five undergraduate linguistic students were recruited for the annotation of both parts. The annotators were provided with written annotation guidelines supplied with examples. 
As a result, we obtained five annotations for one half of the corpus and three annotations for the second half as two annotators annotating the second half were not able to complete their work. The corpus was word-tokenized using the EstNLTK tokenizer. The tokenization was manually checked and corrected by the second author of this paper.

The inter-annotator agreements of the annotations in terms of both Dice coefficient and Fleiss kappa are reported in Table~\ref{tab:agreement}. As can be seen from the table, both Dice and kappa values refer to very high inter-annotator agreement. The agreement is the highest for orthographic sentences and the lowest for syntactic sentences. This is expected as annotating the syntactic sentence boundaries assumes more subjective judgement compared to orthographic boundaries.

\begin{table}[t]
    \caption{Inter-annotator agreements of sentence boundary annotations. Binary boundary refers to the presence or absence of any boundary annotation, regardless whether it is orthographic or syntactic}    \centering
    \begin{tabular}{lcc}
        \toprule
         & \bf Dice & \bf Fleiss $\kappa$ \\
         \midrule
        Binary boundary & 0.92 & 0.91 \\
        Orthographic boundary & 0.96 & 0.95\\
        Syntactic boundary & 0.90 & 0.89\\
        \bottomrule
    \end{tabular}
    \label{tab:agreement}
\end{table}

\section{Sentence Segmentation and Tokenization Evaluation}

For further evaluations, we constructed a corpus containing majority annotations. The decision to retain the boundary was done for both types of sentence boundaries separately---the orthographic boundary was retained in places where at least three (or, in the second half of the corpus, two, respectively) annotators had annotated the orthographic sentence boundary. Similarly, syntactic boundaries were retained in places where at least three (or, in the second half, two) annotators had annotated the syntactic boundary. As a result, the corpus with majority annotations contains three types of sentence boundaries: the majority of annotators had annotated 1) both orthographic and syntactic boundary, 2) only orthographic boundary, and 3) only syntactic boundary. 

We used this corpus to evaluate the performance of three tokenization and sentence segmentation systems: EstNLTK version 1.6.5 \cite{laur2020estnltk}, Stanza \cite{qi2020stanza} and UDPipe \cite{straka2016udpipe}. Est\-NLTK tokenizer is rule-based and the sentence segmenter is based on the Punkt system \cite{kiss2006unsupervised} with additional post-processing rules. For Stanza and UDPipe we used the models pretrained on Estonian UD corpus available on their respective web sites.\footnote{\url{https://stanfordnlp.github.io/stanza/available_models.html}}\footnote{\url{http://ufal.mff.cuni.cz/udpipe/models}}

We compared all systems in three evaluation scenarios:
\begin{enumerate}
    \item \emph{All boundaries}: both orthographic and syntactic boundaries are considered as gold sentence boundaries;
    \item \emph{Orthographic boundaries}: only orthographic boundaries are considered as gold sentence boundaries;
    \item \emph{Relaxed boundaries}: both orthographic and syntactic boundaries are considered as gold sentence boundaries but the system was not penalized when it did not mark the syntactic boundary.
\end{enumerate}
Table~\ref{tab:eval} shows the precision, recall and F1-scores of both sentence segmentation and tokenization results computed with the official evaluation script of the CoNLL 2018 Shared Task \cite{zeman2018conll}, with necessary modifications made for the relaxed boundary evaluation. 

In all evaluation scenarios, the EstNLTK segmentation system performed the best while Stanza and UDPipe perform similarly, with Stanza being slightly worse. The best F-score of orthographic boundaries on this corpus (87.58) is considerably lower than 92.87, which was the best sentence segmentation score reported in the CoNLL 2018 Shared Task on the UD Estonian test.\footnote{\url{https://universaldependencies.org/conll18/results-sentences.html}} When considering both orthographic and syntactic boundaries as gold boundaries, the performance is considerably lower, confirming that existing systems are oriented to detecting orthographic sentence boundaries. 

In terms of tokenization, again the EstNLTK system performs the best and here, also the Stanza and UDPipe systems perform roughly the same. While the tokenization results are very high on all systems, they still remain below the best results reported in the CoNLL 2018 Shared Task, where the best systems achieved an F-score of 99.96 on the Estonian UD test set.

\begin{table}[t]
    \caption{Sentence segmentation and tokenization evaluation results. Orth. stands for orthographic}    \centering
    \begin{tabular}{lccccccccc}
        \toprule
        & \multicolumn{3}{c}{\bf EstNLTK} & \multicolumn{3}{c}{\bf Stanza} & \multicolumn{3}{c}{\bf UDPipe} \\
        & \bf Prec & \bf Rec & \bf F1 & \bf Prec & \bf Rec & \bf F1 & \bf Prec & \bf Rec & \bf F1 \\
        \midrule
        All boundaries & \bf 80.11 & \bf 68.91 & \bf 74.09 & 77.12 & 68.25 & 72.41 & 79.24 & 66.89 & 72.54 \\
        Orth. boundaries & \bf 88.40 & \bf 86.77 & \bf 87.58 & 83.64 & 84.46 & 84.05 & 86.80 & 83.59 & 85.17 \\
        Relaxed boundaries & \bf 88.40 & \bf 85.05 & \bf 86.70 & 84.73 & 83.30 & 84.00 & 87.58 & 83.29 & 85.38\\
        \midrule
        Tokenization & \bf 98.13 & \bf 98.93 & \bf 98.53 & 96.68 & 96.89 & 96.78 & 97.03 & 96.16 & 96.59 \\
        \bottomrule
    \end{tabular}

    \label{tab:eval}
\end{table}

\subsection{Sentence Segmentation Comparison on the UD and EWT Corpora}

To assess how much the out of domain characteristics of the web texts affect the performance of different sentence segmentation systems, we next present in Table~\ref{tab:ud_eval} the sentence segmentation and tokenization results on Estonian Universal Dependencies v2.5 test set. Note that both UDPipe and Stanza systems have been trained on the Estonian UD training set and thus one can expect these systems to perform much better in this scenario. 
For EWT, the orthographic sentence boundary results have been copied from Table~\ref{tab:eval}.

Table~\ref{tab:ud_eval} shows that Stanza obtains the best sentence segmentation results on the more well-formed UD test set, while EstNLTK performs considerably worse than both Stanza and UDPipe. The Stanza performance is also better than the best sentence segmentation result reported on the CoNLL 2018 Shared Task (92.87), which might be due to two reasons: 1) the tokenizer that was part of the Stanford parsing pipeline and now repackaged as the Stanza system has been improved, and 2) the CoNLL 2018 Shared Task systems were trained and evaluated on an older version of the Estonian UD corpus and the performance differences stem from the differences in the different versions of the corpus.

Overall, these results confirm that both rule-based and model-based systems are vulnerable to textual domain characteristics. While the EstNLTK seems to be more biased towards noisy web domain, the supervised model-based Stanza and UDPipe perform much better on the well-formed domain they were trained at. This leads to a quite an obvious suggestion that the performance of the Stanza and UDPipe systems on web texts might be improved if they would be trained on the data consisting of annotated web texts.

\begin{table}[t]
	\caption{Comparison of sentence segmentation systems on the Estonian UD test set and the reannotated EWT corpus with and without paragraph boundaries}
    \centering
    \begin{tabular}{lcccccc|ccc}
        \toprule
        & \multicolumn{3}{c}{\bf Estonian UD} & \multicolumn{3}{c|}{\bf EWT} & \multicolumn{3}{c}{\bf EWT w/o paragraphs}\\
        & \bf Prec & \bf Rec & \bf F1 & \bf Prec & \bf Rec & \bf F1 & \bf Prec & \bf Rec & \bf F1\\
        \midrule
        EstNLTK & 88.60 & 82.92 & 85.66 & \bf 88.40 & \bf 86.77 & \bf 87.58 & \bf 83.64 & 74.70 & \bf 78.91\\
        Stanza & \bf 94.77 & \bf 91.91 & \bf 93.32 & 83.64 & 84.46 & 84.05 & 81.89 & \bf 76.03 & 78.85\\
        UDPipe & 93.40 & 89.79 & 91.56 & 86.80 & 83.59 & 85.17 & 81.43 & 72.21 & 76.55\\
        \bottomrule
    \end{tabular}
    \label{tab:ud_eval}
\end{table}

\subsection{The Effect of Paragraph Boundaries}
If the text contains paragraph boundaries, then sentence segmentation systems get one sentence boundary per paragraph for free. Therefore, maintaining paragraph boundaries in the input text has potentially large effect on sentence segmentation accuracy. 
Unfortunately, on the Estonian UD datasets, the paragraph boundaries have not been retained and thus, systems evaluated on the UD test set cannot make use of this so-called free lunch.

As explained above in Section~\ref{sec:corpus}, the EWT dataset did not initially contain paragraph boundaries, but these were reintroduced when preparing the corpus for sentence boundary annotation. Thus, we can evaluate the effect the presence or absence of paragraph boundaries has to different sentence segmentation systems.
The right-most section of the Table~\ref{tab:ud_eval} shows these results. While the performance of all systems degrade when paragraph boundaries are not available, the EstNLTK scores decrease the most, suggesting that it is most sensitive to the missing paragraph boundaries. This can be also a partial explanation of why the EstNLTK achieves the lowest results on UD test set that does not contain paragraph boundaries.

\section{Error Analysis}
To get an idea about where the systems struggle the most on the web corpus, we manually analyzed and categorized the orthographic sentence boundary errors made by the systems. The common error types and the proportional division of errors into these types of different systems are given in Table~\ref{tab:errors}. The overall pattern of error division is roughly similar for Stanza and UDPipe and somewhat different for EstNLTK.

A large category of errors for all systems consists of missing boundaries after multiple punctuation marks (1), typically three dots (...). The complement to these errors is the category 6: wrongly placed boundary after multiple punctuation marks.
These situations are often ambiguous also to the human annotators and can be interpreted either as a sentence boundary or a pause in the middle of the sentence, depending on the context and the perception of the individual annotator.

Another complementary pair of error categories is no boundary due to the missing sentence-final punctuation (3) and boundary in the middle of a sentence (5). The latter error often occurs when the next word starts with a capital letter (like a name) and so the systems erroneously decide that this must be the first word of the next sentence. This error only occurs with Stanza and UDPipe and was never observed with EstNLTK. EstNLTK, on the other hand, makes more of the errors of category 3 where the sentence boundary is missed due to the absence of the sentence final punctuation in the text.

Other error categories are mostly due to incorrect tokenization. For instance, both Stanza and UDPipe fail to occasionally put a sentence boundary after a valid sentence final punctuation mark (2). These errors occur mostly in situations where there is no space character between the punctuation mark and the next word, thus leading these systems to predict the whole sequence as a single token. On the other hand, EstNLTK makes a larger proportion of errors compared to Stanza and UDPipe by placing a sentence boundary inside a token that contains a punctuation mark (8). These problems are again possible due to the tokenization errors. 
Also, abbreviations can confuse the systems. Sometimes the punctuation symbol right after the abbreviation should be part of the abbreviation token and thus, does not denote the sentence boundary and, in some cases, the punctuation mark signifies the end of the sentence (9).
Finally, if the tokenization system splits the repeated punctuation marks (4) or multiple punctuation marks making up emoticons (7) into several tokens, then these can also cause errors in the subsequent sentence segmentation.

\begin{table}[t]
    \caption{The proportion of errors belonging to different categories made by different sentence segmentation systems. The type indicates whether a boundary was missing (M) or an extra boundary was added (A)}
    \centering
    \setlength{\tabcolsep}{5pt}
    \begin{tabular}{rlcrrr}
    \toprule
        \bf No & \bf Error category & \bf Type & \bf EstNLTK & \bf Stanza & \bf UDPipe \\ 
        \midrule
        1 & No boundary after multiple punctuation marks & M & 37 \% & 16 \% & 19 \% \\
        2 & No boundary after sentence final punctuation & M & 0 \% & 19 \% & 34 \% \\
        3 & No boundary due to missing sentence-final punctuation & M & 18 \% & 8 \% & 12 \% \\
        4 & Sentence boundary inside repeated punctuation marks & A & 14 \% & 17 \% & 7 \% \\
        5 & Boundary in the middle of a sentence & A & 0 \% & 20 \% & 12 \% \\
        6 & Wrong boundary after multiple punctuation marks & A & 5 \% & 9 \% & 7 \% \\
        7 & Emoticon has been segmented as a separate sentence & A & 6 \% & 6 \% & 4 \%\\
        8 & Sentence boundary inside a token & A & 12 \% & 2 \% & 1 \% \\
        9 & Missing boundary after sentence-final abbreviation token & M & 3 \% & 2 \% & 2 \% \\
         & Others & & 3 \% & 2 \% & 2 \% \\
        \bottomrule
    \end{tabular}
    \label{tab:errors}
\end{table}

\section{Discussion and Conclusions}

In this paper, we described a sentence boundary annotation effort of the EWT corpus and presented the sentence segmentation and word tokenization results of three segmentation systems on this dataset. These results were compared to the sentence and token segmentation performance of the same systems on the more well-formed UD dataset. We found that on the newly annotated web corpus, the EstNLTK system based on the Punkt model with additional rules performs the best, while on the more well-formed UD test set, the neural models based Stanza and UDPipe systems performed better. Based on these results, our suggestion would be to use EstNLKT system for processing noisy web texts and prefer model-based Stanza or UDPipe systems to segment well-formed written texts. Moreover, the error analysis presented in the previous section indicates that the performance of supervised UDPipe and Stanza systems might be improved on web texts when the training data of these models would also contain web domain data. 

Sentence segmentation is a crucial preprocessing step for syntactic analysis systems, which rely on properly detected sentence boundaries. Although model-based syntactic parsers can deal with long orthographic sentences often found in noisy web texts, we hypothesise that syntactic analysers will perform better if these long orthographic sentences are split into several syntactically independent parts that can be analysed independently. 
The dataset presented in this paper also contains the annotations of syntactic sentence boundaries which enables future work to test out this hypothesis.
That would involve evaluating the accuracy of dependency parsing performane on two versions of this dataset: one containing only orthographic sentence boundaries and another containing both orthographic and syntactic boundaries.

\section*{Acknowledgements}
We thank Kadri Muischnek for the help in developing the annotation guidelines and the students of linguistics who annotated the corpus.

\bibliographystyle{vancouver}
\bibliography{BalticHLT2020_Sirts_Peekman}
\end{document}